# Augmenting Human-Annotated Training Data with Large Language Model Generation and Distillation in Open-Response Assessment

Conrad Borchers[1], Danielle R. Thomas[1], Jionghao Lin[1], Ralph Abboud[2], Kenneth R. Koedinger[1]
[1]Carnegie Mellon University
[2]Learning Engineering Virtual Institute
{cborchers,drthomas,jionghao,koedinger}@cmu.edu
rabboud@levimath.org

**ABSTRACT**: Large Language Models (LLMs) like GPT-4o can help automate text classification tasks at low cost and scale. However, there are major concerns about the validity and reliability of LLM outputs. By contrast, human coding is generally more reliable but expensive to procure at scale. In this study, we propose a hybrid solution to leverage the strengths of both. We combine human-coded data and synthetic LLM-produced data to fine-tune a classical machine learning classifier, distilling both into a smaller BERT model. We evaluate our method on a human-coded test set as a validity measure for LLM output quality. In three experiments, we systematically vary LLM-generated samples' size, variety, and consistency, informed by best practices in LLM tuning. Our findings indicate that augmenting datasets with synthetic samples improves classifier performance, with optimal results achieved at an 80% synthetic to 20% human-coded data ratio. Lower temperature settings of 0.3, corresponding to less variability in LLM generations, produced more stable improvements but also limited model learning from augmented samples. In contrast, higher temperature settings (0.7 and above) introduced greater variability in performance estimates and, at times, lower performance. Hence, LLMs may produce more uniform output that classifiers overfit to earlier or produce more diverse output that runs the risk of deteriorating model performance through information irrelevant to the prediction task. Filtering out inconsistent synthetic samples did not enhance performance. We conclude that integrating human and LLM-generated data to improve text classification models in assessment offers a scalable solution that leverages both the accuracy of human coding and the variety of LLM outputs.

**Keywords**: Data augmentation, GPT-4o, Large Language Models, Open-response assessment, Model distillation, Few-shot prompting, Text classification, Synthetic data generation

## 1 Introduction and Related Work

Large language models (LLMs) have emerged as valuable tools for automating text classification at scale, accelerating research in Learning Analytics and its related fields. Text classification in learning analytics has a variety of applications ranging from assessment of short answer responses, coding dialogue for different dialogue acts, and inferring regulated learning processes. For instance, LLMs have been used to automatically code self-regulated learning processes from think-aloud data during tutoring system practice (Zhang et al., 2024) and facilitate codebook development for human annotation (Barany et al., 2024). LLMs have also demonstrated potential to improve learning systems by classifying dialogue acts in peer tutoring scenarios, another classification process typically done by human annotators (Borchers et al., 2024) and in analyzing feedback during short-answer grading and training human tutors (Lin et al., 2024). While these methods are faster, cheaper, and more scalable than human annotation for open-response assessment tasks (Thomas et al., 2024a), they can also be more unreliable or hallucinate irrelevant or false output (Huang et al., 2023).

Given the importance of accurate and reliable classifiers in educational applications, the field of learning analytics has made efforts in creating and rigorously evaluating LLM classifiers. This emerging body of research has predominantly followed two main approaches to classifying text using LLMs (Carpenter et al., 2024). A first approach to overcoming unreliable LLM output is using their embeddings to train more reliable, classical models on human-coded data (Zhang et al., 2024). This approach enables the creation of offline, stable, and interpretable models but limits the potential of LLMs by relegating them to a feature augmentation role, thereby sacrificing their strong generative abilities. The second approach directly prompts LLMs to grade or classify responses, validating these outputs against human-coded data (Thomas et al., 2024a). However, this approach faces challenges as it requires extensive prompt engineering and occasional inconsistent formatting of responses or suboptimal performance due to insufficient variety in training data featured in prompt instruction.

Both aforementioned approaches require significant amounts of human data as a source of truth, which is labor-intensive. In light of these approaches, we propose to combine LLM-based generation and prediction to address their respective limitations. We use state-of-the-art LLMs to augment training data and improve classical machine learning techniques learning from text. Specifically, we prompt LLMs to generate graded example responses guided by established codebook criteria and instructions. We then augment human-coded training data with these synthetic responses.

Our approaches differ from the aforementioned approaches in two key ways: First, unlike embedding-based LLM classifiers trained on human-coded data, we increase the variety and representativeness of models' training sets, which should lead to performance improvements depending on the LLM's ability to provide high-quality data. Second, unlike direct LLM labelling (i.e., prediction through generation), we use LLM generation as an input feature, enabling us to leverage limited human-coded data to better inform prompting strategies and LLM tuning.

By training a smaller model on both human-coded data and the LLMs generations, we effectively *distill* the knowledge represented within the LLM relevant to the prediction task (Hsieh et al., 2023). Moreover, by training on human-coded data *first* and then observing further performance improvements stemming from adding synthetic samples, we validate the LLM's ability to generate meaningful samples on a human-coded holdout test set. This hybrid approach leverages both the rigor and real-world representativeness of human-coded learner data and the variety and scalability of LLM-generated samples, which may enhance the predictive performance of the resulting classifier.

### 1.1 The Present Study

To rigorously conduct this study, we carefully investigate a set of key design decisions in combining LLM-generated training samples with human-coded data. First, temperature settings control the variety and randomness of LLM output, and has been noted as an important factor in LLM performance (Renze & Guven, 2024). In our case, high output variety may improve a model's ability to learn from diverse examples, but excessive randomness may make generations irrelevant to the prediction task. Second, we study the optimal ratio of human-to-generated samples, as past work has noted that synthetic samples considerably larger than the original training corpus can further boost predictive accuracy (Yuan et al., 2023). To address potentially inconsistent output and hallucinations (Huang et al., 2023), we also study the utility of post-processing generated data to remove inconsistencies. Specifically, we filtered inconsistent generations through prompting the LLM to evaluate its own output. This experiment is motivated by research on self-consistency, demonstrating that LLM performance on reasoning tasks can be enhanced by following the most probable generation based on the same prompt (Wang et al., 2023). We study the following research questions:

RQ1: How does predictive performance change based on the human-coded testing set as a function of the quantity of human-coded and synthetic training data?

RQ2: What is the optimal temperature for generating performance-improving synthetic samples?

RQ3: How effective is post-processing based on LLM consistency for distillation effectiveness?

## 2 METHODS

We leveraged learner open-response data from an online lesson, where college-student tutors practice responding to tutoring scenarios involving middle-school math students (Thomas et al., 2024b. Learners produced observations (X, y) where X corresponds to their open-ended response to lesson questions and y to the related grade label (0 if correct or 1 if incorrect), which were used in a prediction task fine-tuning a BERT classifier, alongside synthetic observations (Figure 1).

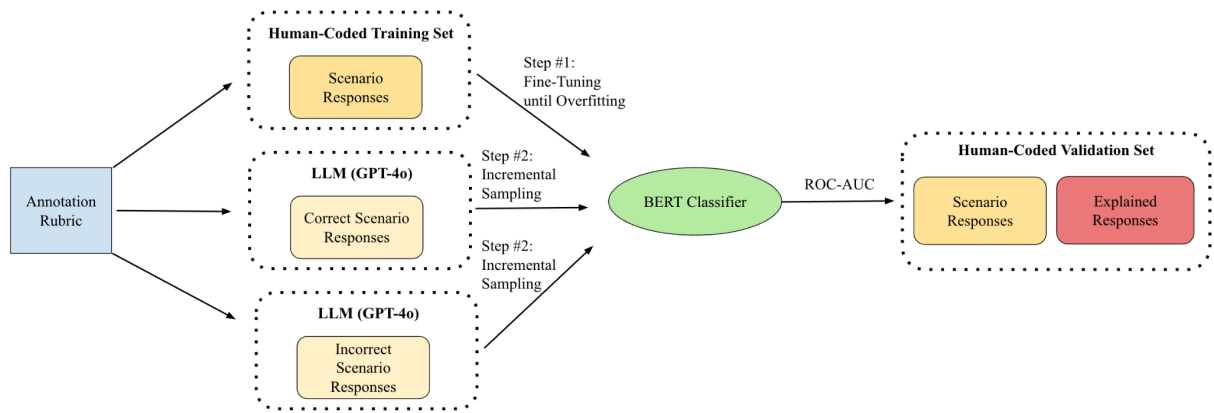

Figure 1: Training process schema. Step 1 illustrates fine-tuning on human-coded data until saturation and Step 2 shows data augmentation using incremental synthetic sampling.

### 2.1 Dataset and Lesson Context

The lesson focused on building tutors' cultural competence—recognizing, valuing, and incorporating students' diverse cultural identities and perspectives into tutoring. Figure 2 illustrates a tutoring scenario tasking tutors to *predict* the best response to a student to increase the tutor's cultural understanding and best support the student. Learner-sourced correct responses to the scenario in Fig. 2 include: *"Marcelo, I'd love to learn more about you and your experiences. What are some things you miss about Mexico or find different here in the U.S.?"* and *"Hello Marcelo! I heard you recently moved from Mexico. I've always been very curious about the country. Would you like to share some of your experiences?* Tutors were also required to *explain* why they thought their response was appropriate, referred to as *explained* responses (see Figure 1).

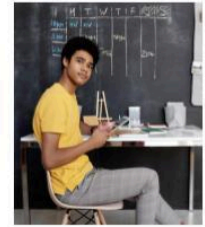

**Figure 2: Sample scenario from the online lesson prompting tutors to *predict* the best approach.**

## 2.2  Human Annotation Process

Two human coders annotated tutor responses according to a rubric.[1] The coders coded responses of 81 tutors and established inter-rater reliability of Cohen's κ = 0.75 and 0.74 for *predicted* and *explained* responses, respectively, which is satisfactory (McHugh, 2012). Based on the human-coded responses, we defined a fine-tuning set of 51 *predicted* responses and a validation set of 256 responses, including both *predict* (102) and *explain* responses (154) for the pre- and post-tests. All explained responses were added to the validation set as a form of validation and generalization task, as they are expected to follow similar but distinct distributions. The first scenario is a pretest, followed by tutors observing the research-recommended approach and receiving feedback. Then, tutors engage in an analogous second scenario as a posttest. The train and test set size were chosen so that performance improvements on the evaluation set could be most reliably estimated, using bootstrapped confidence intervals of performance based on ROC AUC, using 1,000 samples.

## 2.3  Prediction Task and Model Architecture

The objective of the classification task is to consider a learner's text input and predict whether the response was appropriate or not based on the coding rubric. We used a fine-tuning approach, adjusting the weights of a pre-trained BERT classifier based on a few examples. We used the BERT-base-uncased model for offline fine-tuning with an ADAM optimizer and a learning rate of 2e-5.

## 2.4  Establishing a Baseline Model Based on Human-Coded Data Only

We trained a baseline classifier based on 51 training samples coded by humans. We fine-tuned the BERT model based on multiple epochs with a patience of two, meaning that the model training terminated if no more validation accuracy improvements were observed after two epochs.

## 2.5  Data Augmentation and Generation Process

To generate additional training data with the GPT-4o LLM (`gpt-4o-2024-08-06`), we created a few-shot prompt with human-coded examples and a rubric to ask the model to create 10 example responses per prompt, divided by new lines. The prompt engineering approach involved designing two distinct systems and user prompts: one to elicit positive, culturally sensitive responses and another to elicit neutral or culturally unaware responses. The former was assigned the outcome label 1, while the latter was assigned 0. For the prompt of generating correct responses, examples emphasized responses that made students feel valued, noticed, and connected by referencing their cultural background or expressing interest in their identity (e.g., asking about cultural practices or drawing parallels to historical contributions). For the prompt of generating undesired responses, examples intentionally lacked such connections, offering neutral or generic feedback focused solely on the task without any cultural engagement. The prompts provided specific examples and a structured format to guide the model in generating responses aligned with each category (see supplementary GitHub repository for full prompt details).

---

[1] https://github.com/conradborchers/bert-distillation

## 2.6 Validation of Synthetic Samples

We validated augmented samples for each temperature setting (0.3, 0.5, 0.7, and 1) using GPT-generated gradings of its own generations based on the same rubric. First, we randomly sampled 100 generated instances from both desired (1) and undesired (0) responses in the dataset per temperature setting, ensuring equal representation and minimizing class imbalance. Then, GPT-generated responses were scored by prompting the GPT model using the same evaluation rubric provided during response generation. GPT was tasked to output a 1 if the output was adequate and a 0 if inadequate. Output was then parsed using a function that parsed JSON outputs and extracted the "Score" attribute, aligning outputs with binary categories (0 and 1).

We calculated several metrics, including Cohen's κ, accuracy, precision, recall, and F1 score, to study model consistency performance across temperature settings. Inconsistent generations, or in other words, generations that were supposed to be generated to be of a certain category but then classified by the same LLM as another, were filtered out in one of our evaluation experiments. The validation results demonstrated high consistency across temperature settings, with Cohen's κ ranging from 0.83 to 0.92, accuracy between 92% and 96%, and F1 scores from 0.91 to 0.96. Precision remained consistently at 1.0, while recall varied slightly, from 0.83 to 0.92. These metrics indicate robust consistency, with minor recall trade-offs as temperature increases, meaning the LLM became slightly more conservative in grading its own generations as temperature increases.

## 2.7 Model Evaluation and LLM Distillation Experiments

We ran a set of evaluation experiments to study if augmented samples generated by the LLM can improve baseline performance beyond the model trained solely on human labels (see Section 2.4). After establishing the baseline accuracy on the validation side using the early stopping rules on human data, we continued training the model using synthetic samples, randomly sampling synthetic samples, and introducing them in increments of 25 observations. We bootstrapped 95% confidence intervals for the AUC performance on the validation set with 1,000 resamples for all experiments.

**Experiment 1:** To investigate the impact of the ratio of human to synthetic samples on predictive accuracy (RQ1), we utilized the GPT-4o model with a fixed temperature setting of 0.3. Synthetic samples were incrementally added until the dataset comprised 250 synthetic samples, resulting in approximately 85% of the training corpus being synthetically generated. This upper bound allowed us to assess performance in a scenario heavily reliant on synthetic data.

**Experiment 2:** To determine the optimal temperature setting concerning predictive accuracy (RQ2), we replicated Experiment 1 using temperature settings of 0.3, 0.5, 0.7, and 1. Higher temperature settings are expected to yield more varied and diverse responses from the LLM. This experiment examined whether increased variety in synthetic samples enhances performance by mitigating early overfitting on similar examples. Conversely, excessively diverse samples might deviate from relevant information within the model's knowledge space, potentially degrading predictive accuracy.

**Experiment 3:** We replicated Experiment 1 using the best-performing temperature setting identified in Experiment 2, this time removing training samples that were inconsistently generated by the language model (see LLM generation validation above). This experiment addressed whether excluding inconsistent examples further improves model performance (RQ3).

## 3    RESULTS

### 3.1    RQ1: Predictive Performance by Data Augmentation Ratio

Predictive performance improved when adding synthetic samples, based on increased AUC performance after stopping model training on human data (Figure 3). The degree of improvement varies depending on the ratio of synthetic samples to human-labeled samples. Across temperature settings, the AUC scores tended to increase slightly with higher ratios of synthetic samples up to a certain point, peaking around 200 synthetic samples, or about a ratio of 80% synthetic to 20% human-coded data (Figure 3). Additional augmentation, similar to continuing to train on the same human-coded data, did not lead to further improvements, indicating overfitting on the training data, as was observed for the human-coded data. Taken together, the LLM generations increased signal and variability in training samples but were exhausted after continued training, indicating overfitting after a certain threshold. Notably, there was also a trend whereby the higher the temperature setting was, the longer the model improved with more samples added to the training corpus.

### 3.2    RQ2: Predictive Performance by Temperature Setting for Data Generation

The plots for different temperature settings (0.3, 0.5, 0.7, and 1) show that lower temperatures (0.3 and 0.5) result in relatively more stable AUC scores with narrower confidence intervals, indicating more consistent improvements in predictive performance (Fig. 3a-b). Higher temperatures (0.7 and 1) introduce greater variability in performance, as evidenced by somewhat wider confidence intervals around the AUC estimate, although they occasionally lead to slightly higher maximum AUC scores (Fig. 3c-d). In this case, performance gains also seem to occur later, indicating a tradeoff between quality and variability. Overall, lower temperature led to more stable performance improvements, but overfit faster (limiting variability and performance improvement) while higher temperature led to less stable performance improvements, but overfit later, overall learning and boosting performance more.

In this experiment, the optimal temperature setting for generating synthetic samples was 0.5, with the best AUC of 0.774 at 200 added samples (Fig. 3b, red arrow). However, based on inspection of 95% confidence intervals, model improvements were likely not statistically significant, though tangible over the 0.744 baseline. Statistical power is limited by the evaluation sample size.

### 3.3    RQ3: Predictive Performance when Removing Inconsistent Samples

For a temperature setting of 0.5, we retained a total of 956 out of the 1,000 samples (96%) And re-run our model evaluation. Notably, performance peaked after only 100 augmented sample sizes, rather than 200, but performance was *not* better than in previous experiments when removing inconsistent samples (AUC = 0.766 compared to AUC = 0.774 in Experiment 2).

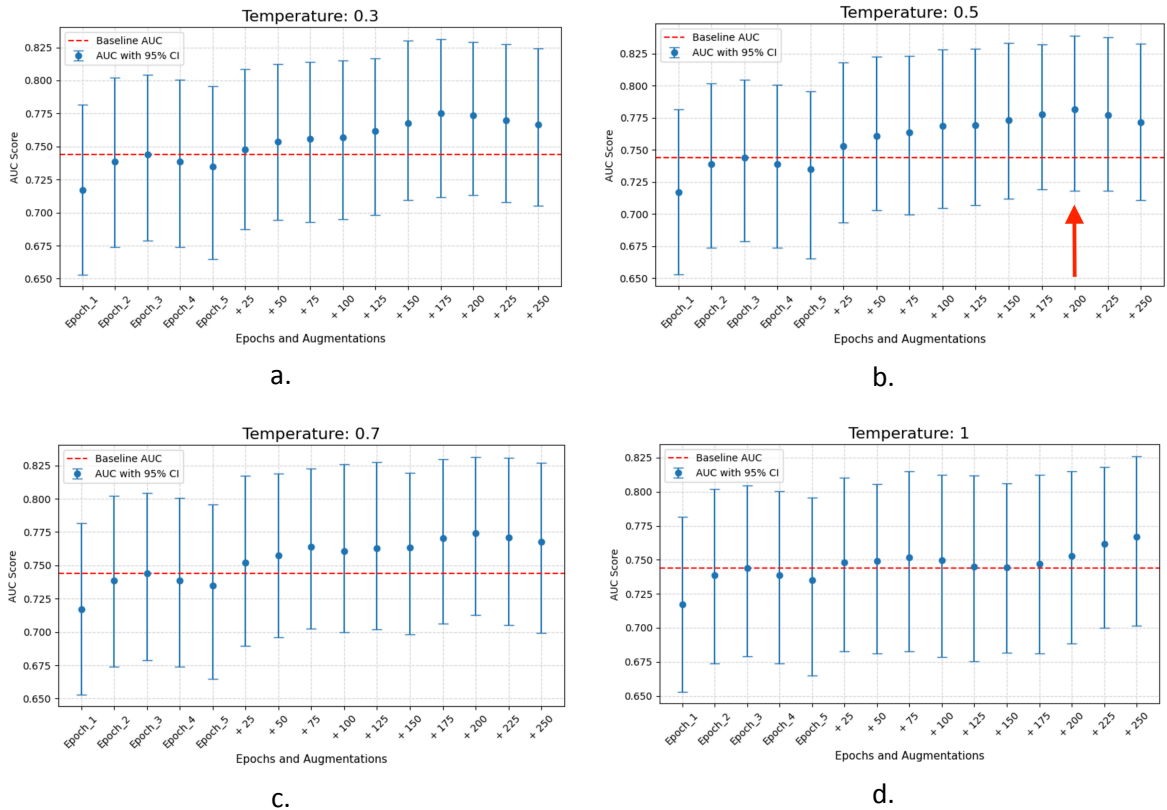

Figure 3: AUC performance with synthetic samples at temperatures 0.3, 0.5, 0.7, and 1 (a-d). The best AUC (0.774) occurs in Fig. 3b at 0.5 temperature with 200 synthetic samples (red arrow).

## 4 Discussion

To address the limitations of relying solely on human-coded data or direct LLM prompting for text classification in assessments, we combined both approaches by augmenting human-coded datasets with GPT-4o-generated synthetic samples. By prompting the LLM to produce culturally responsive and non-responsive tutor responses based on rubrics, we investigated how varying the ratio of synthetic to human-labeled data and controlling sample variety through temperature settings could enhance open-response assessment accuracy, a common task in learning analytics and education.

### 4.1 Few-Shot Prompting for Synthetic Data Generation Improves Predictive Performance Beyond Human-Coded Data, but Requires Regularization

Building upon a baseline established with human-coded responses from real learner data, we demonstrated that augmenting the training corpus with GPT-4o-generated synthetic samples can enhance the predictive accuracy of a distilled BERT classifier on a holdout test set. Including few-shot prompted synthetic data allowed the model to capture a broader range of response patterns and culturally responsive interactions that may not have been fully represented in the limited human-coded dataset of real learners' responses used for fine-tuning. Although our sample size was not large enough to confirm that specific improvements were statistically significant, the consistent trend across multiple independent experiments indicates that incorporating LLM-generated data enriches the model's ability to recognize relevant real-world knowledge.

This suggests that large language models can effectively widen the signal for open-response assessment tasks by utilizing their vast knowledge base (being trained on open web data) beyond responses generated by our human-coded sample. One potential reason is that our study sample of human-coded data constitutes responses by a specific learner population in a specific culture at a particular time, that is, American college students learning to become tutors. In contrast, the LLM may be able to generate responses that are not found in our sample but are still relevant to the prediction task, hence improving model accuracy and its ability to generalize to new observations.

Our findings merit additional experiments to improve open-response grading through this novel data augmentation method. Our method may be combined with other augmentation methods that exist in the literature, such as linguistic preprocessing methods and sample duplications (see Wei & Zou, 2019), which may be combined with the methods demonstrated here in future research.

## 4.2 Reducing Variance in Generation Through Temperature Requires Fewer Samples Until Models Start Overfitting During Training, But May Limit Performance

Across Experiments 2 and 3, we observed a general trend where improvements in model performance achieved through GPT-4o-augmented data tended to plateau earlier when the generated responses were less varied. This reduction in variation was observed for (a) lower temperature settings in Experiment 2 and (b) by removing inconsistent samples in Experiment 3. This suggests that more varied responses of GPT-4o may also generate signals that the model can effectively learn to improve performance on the task, meaning that it can learn for longer and from more training samples without overfitting, as indicated by a plateau or reduction in performance on the validation set upon further training. Interestingly, in the latter case, removing inconsistent samples—which might represent rare or unique response types—counterintuitively led to a decrease in performance. This suggests that these inconsistent samples may cover a broader spectrum of the response space valuable for the model's learning. It could be that these samples are also specifically difficult to classify, pointing to potential refinements to our codebook used by human coders, which merits further investigation and future research. This research could expand on nascent human–AI collaborative codebook development investigations in qualitative research (Barany et al., 2024).

In all experiments, our results suggest that some form of regularization is necessary to control the extent of variation in the synthetic data generated by the LLM to avoid overfitting to noise and ensure generalizability. However, our experiments highlight that models have the potential to learn more effectively from datasets that include a higher variety of responses, provided they are of sufficient quality. This increased variety may improve classifier performance, but also necessitate more training data to enable the model to distinguish between varying response qualities effectively. Overall, our results suggest that LLMs are a high-quality data source for training open-response assessment models if used carefully. While excessive variety can decrease performance, the LLM's output variety was constructive concerning our target evaluation.

In sum, our findings suggest that methods to study variation and quality of LLM-generated training data have the potential to further extend performance improvements documented in the present study. Avenues for further refinement include fine-tuning and regularizing variation in GPT-generated training samples. Techniques such as quantifying similarity through cosine similarity, clustering responses, or employing advanced prompt engineering strategies to elicit varied responses could be explored. Additionally, oversampling rare response types might enhance the model's ability to generalize from diverse data, improving predictive performance in complex assessment tasks.

### 4.3 Limitations and Future Research

First, we only employed few-shot prompting for synthetic data generation to augment our human-coded training corpus. Our prompts may have yet to fully optimize synthetic data's variety and relevance through better instructions, and perhaps more diverse examples (as we used the same few-shot examples for all generated data). Future research may explore advanced prompt techniques to elicit more varied and representative samples. Improved generation protocols may also require less human-coded data to reach performance saturation, another direction for future research.

Second, we generated fully synthetic training pairs consisting of responses and labels. Future work may also explore the use of GPT to generate labels for existing unlabeled data as a data augmentation strategy. We previously conducted an experiment where GPT-4o generated labels for uncoded learner inputs. However, our data were limited to 50 samples of real, uncoded learner responses. As our results suggest that more samples are needed to maximize predictive performance, we decided not to report findings from this experiment in the present manuscript.

Third, we used only one generative model (GPT-4o) and distillation architecture (BERT). Future research may attempt to replicate our experiments with other LLMs, including open-source LLMs, and other distillation architectures, such as simple logistic regression models (or more complex models as classical models like BERT may limit performance). Similarly, our focus on a single context of building cultural awareness limits the generalizability of our findings. Successful replication would provide further evidence that LLMs can be used to augment human data in assessment tasks.

### 5 CONCLUSION

Our study demonstrates that augmenting human-coded datasets with GPT-4o-generated synthetic samples can tangibly improve classifier performance in open-response grading. Managing the variety introduced by synthetic data requires effective regularization. While greater variation can enrich model learning and improve performance, it also risks generating irrelevant samples to the task, leading to worse performance. Therefore, in general, the potential of GPT for data augmentation becomes larger with more diverse sampling but also riskier in selecting error island examples, making post-processing, which we only explored on the surface, even more critical alongside other regularization methods. Taken together, our results highlight the effectiveness of LLMs as a high-quality data source to augment human-coded data under the right circumstances.

We chart several directions that merit future research. For example, we still need to explore methods to control similarity among synthetic samples fully. In future work, we plan to utilize sentence embeddings and cosine similarity measures to balance variety and relevance, potentially further enhancing model performance beyond the model performance improvements documented here.

ACKNOWLEDGEMENTS

This work was made possible with the support of the Learning Engineering Virtual Institute. Any opinions, findings, and conclusions expressed in this material are those of the authors.

## REFERENCES

Barany, A., Nasiar, N., Porter, C., Zambrano, A. F., Andres, A. L., Bright, D., ... & Baker, R. S. (2024). ChatGPT for education research: exploring the potential of large language models for qualitative codebook development. In *International Conference on Artificial Intelligence in Education* (pp. 134-149). Cham: Springer Nature Switzerland.

Borchers, C., Yang, K., Lin, J., Rummel, N., Koedinger, K. R., & Aleven, V. (2024). Combining dialog acts and skill modeling: What chat interactions enhance learning rates during AI-supported peer tutoring? *Proceedings of the 17th International Conference on Educational Data Mining.*

Carpenter, D., Min, W., Lee, S., Ozogul, G., Zheng, X., & Lester, J. (2024). Assessing student explanations with large language models using fine-tuning and few-shot learning. In *19th Workshop on Innovative Use of NLP for Building Educational Applications* (BEA 2024).

Huang, L., Yu, W., Ma, W., Zhong, W., Feng, Z., Wang, H., ... & Liu, T. (2023). A survey on hallucination in large language models: Principles, taxonomy, challenges, and open questions. ACM *Transactions on Information Systems.*

Hsieh, C. Y., Li, C. L., Yeh, C. K., Nakhost, H., Fujii, Y., Ratner, A., ... & Pfister, T. (2023). Distilling step-by-step! Outperforming larger language models with less training data and smaller model sizes. arXiv preprint arXiv:2305.02301.

Lin, J., Han, Z., Thomas, D. R., Gurung, A., Gupta, S., Aleven, V., & Koedinger, K. R. (2024). How can I get it right? Using GPT to rephrase incorrect trainee responses. *International Journal of Artificial Intelligence in Education*, 1-27.

McHugh, M. L. (2012). Interrater Reliability: The Kappa Statistic. Biochemia Medica, 22(3), 276-282.

Renze, M., & Guven, E. (2024). The effect of sampling temperature on problem solving in large language models. arXiv preprint arXiv:2402.05201.

Thomas, D. R., Borchers, C., Kakarla, S., Lin, J., Bhushan, S., Guo, B., ... & Koedinger, K. R. (2024a). Do Tutors Learn from Equity Training and Can Generative AI Assess It?. arXiv preprint arXiv:2412.11255.

Thomas, D. R., Borchers, C., Kakarla, S., Lin, J., Bhushan, S., Guo, B., ... & Koedinger, K. R. (2024b). Does Multiple Choice Have a Future in the Age of Generative AI? A Posttest-only RCT. arXiv preprint arXiv:2412.10267.

Wang, X., Wei, J., Schuurmans, D., Le, Q. V., Chi, E. H., Narang, S., ... & Zhou, D. (2023) Self-Consistency Improves Chain of Thought Reasoning in Language Models. In The Eleventh International Conference on Learning Representations.

Wei, J., & Zou, K. (2019). Eda: Easy data augmentation techniques for boosting performance on text classification tasks. arXiv preprint arXiv:1901.11196.

Yuan, J., Zhang, J., Sun, S., Torr, P., & Zhao, B. (2023) Real-fake: Effective training data synthesis through distribution matching. *12th International Conference on Learning Representations.*

Zhang, J., Borchers, C., Aleven, V., & Baker, R. S. (2024). Using large language models to detect self-regulated learning in think-aloud protocols. In *Proceedings of the 17th International Conference on Educational Data Mining (EDM).*